\documentclass[letterpaper, 10 pt, conference]{ieeeconf} 
                                                         
\IEEEoverridecommandlockouts

\usepackage{flushend}
\usepackage{balance} 

\let\labelindent\relax

\usepackage{etoolbox} 
\usepackage{times}
\usepackage[utf8]{inputenc} 
\usepackage[T1]{fontenc}    
\usepackage[english]{babel} 
\usepackage[protrusion=true,expansion=true]{microtype}
\usepackage{comment}
\usepackage{cancel}
\usepackage{csquotes}
\usepackage{listings}
\usepackage{nameref}
\usepackage{paralist}
\usepackage{xspace}
\usepackage{setspace}
\usepackage{makeidx}
\usepackage{pdfpages}

\usepackage{amsmath,amssymb} 
\usepackage{amsfonts}       
\usepackage{mathtools}
\usepackage{array} 
\usepackage{nicefrac}       
\usepackage{breqn}          
\usepackage{textcomp}
\usepackage{bm} 
\usepackage{pifont}
\usepackage[
  separate-uncertainty = true,
  multi-part-units = repeat
]{siunitx}

\usepackage{graphicx}
\usepackage{adjustbox} 
\usepackage{epsfig}

\usepackage{float}
\usepackage[font=small]{subcaption}
\usepackage[font=small,labelfont=bf]{caption}
\usepackage{lscape}      
\usepackage{wrapfig}

\usepackage{tikz}  
\usepackage{makecell}
\usepackage{overpic}
\usepackage{algorithm}
\usepackage[noend]{algpseudocode}

\usepackage{array} 
\usepackage{tabularx}
\usepackage{multirow}
\usepackage{multicol}
\usepackage{booktabs}   

\usepackage{paralist}
\usepackage{enumitem}
\setlist[itemize]{noitemsep,leftmargin=*,topsep=0in}
\setlist[enumerate]{noitemsep,leftmargin=*,topsep=0in}

\usepackage{url}            
\PassOptionsToPackage{table}{xcolor}
\usepackage{color,colortbl} 
\usepackage{soul}

\PassOptionsToPackage{capitalize, noabbrev}{cleveref}
\usepackage[pagebackref=false,breaklinks=true,colorlinks,urlcolor=blue,citecolor=blue,linkcolor=blue,bookmarks=true]{hyperref}

\makeatletter
\let\NAT@parse\undefined
\makeatother
\usepackage[numbers,sort&compress]{natbib}

\usepackage{blindtext}
\usepackage{bbm}

\usepackage{lipsum}

\newcommand{\includegraphicsmaybe}[1]{\IfFileExists{#1}{\includegraphics{#1}}{\includegraphics{dummy.pdf}}}

\usepackage{titlesec}
\titlespacing{\section}{0pt}{0.3\baselineskip}{0.25\baselineskip}
\titlespacing{\subsection}{0pt}{0.25\baselineskip}{0.15\baselineskip}
\titlespacing{\subsubsection}{0pt}{0.05\baselineskip}{0.03\baselineskip}

\renewcommand{\paragraph}[1]{\vspace{0.2em}\noindent\textit{#1} --}

\makeatletter
\newcolumntype{B}[3]{>{\boldmath\DC@{#1}{#2}{#3}}c<{\DC@end}}
\makeatother

\graphicspath{ {figures/} }

\usepackage[normalem]{ulem}

\definecolor{codegreen}{rgb}{0,0.6,0}
\definecolor{codegray}{rgb}{0.4,0.4,0.4}
\definecolor{codepurple}{rgb}{0.5,0,0.9}
\definecolor{backcolour}{rgb}{0.95,0.95,0.95}

\lstdefinestyle{mystyle}{
    backgroundcolor=\color{backcolour},   
    commentstyle=\color{codegreen},
    keywordstyle=\color{magenta},
    numberstyle=\tiny\color{codegray},
    stringstyle=\color{codepurple},
    basicstyle=\fontsize{6.5}{7}\selectfont\ttfamily\ttfamily,
    breakatwhitespace=false,         
    breaklines=true,
    breakindent=0pt,
    captionpos=b,                    
    keepspaces=true,                 
    numbers=none,                    
    numbersep=5pt,                  
    showspaces=false,                
    showstringspaces=false,
    showtabs=false,                  
    tabsize=2,
}
\renewcommand{\paragraph}[1]{\vspace{0.1em}\noindent\textit{#1} --}

\newcolumntype{C}[1]{>{\centering\let\newline\\\arraybackslash\hspace{0pt}}m{#1}}

\usepackage{xspace}
\def\ourmodel{\textsc{CLAIRify}\xspace}

\title{\LARGE \bf
Errors are Useful Prompts: Instruction Guided Task Programming \\ with Verifier-Assisted Iterative Prompting
}

\author{Marta Skreta$^{\ast1,2}$,
        Naruki Yoshikawa$^{\ast1,2}$,
        Sebastian Arellano-Rubach$^{3}$,
        Zhi Ji$^{1}$,
        Lasse Bjørn Kristensen$^{1}$, \\
        Kourosh Darvish$^{1,2}$,
        Al\'{a}n Aspuru-Guzik$^{1,2}$,
        Florian Shkurti$^{1,2}$,
        Animesh Garg$^{1,2,4}$
\thanks{$^\ast$ Authors contributed equally,
    $^{1}$University of Toronto,
    $^{2}$Vector Institute,
    $^{3}$University of Toronto Schools,
    $^{4}$NVIDIA}
\thanks{Email: \texttt{\{martaskreta,naruki\}@cs.toronto.edu}}
\thanks{Email: \texttt{\{kdarvish,garg\}@cs.toronto.edu}}
}

\begin{document}
\bstctlcite{IEEEexample:BSTcontrol}
\maketitle
\thispagestyle{empty}
\pagestyle{empty}

\begin{abstract}
Generating low-level robot task plans from high-level natural language instructions remains a challenging problem. Although large language models have shown promising results in generating plans, the accuracy of the output remains unverified. Furthermore, the lack of domain-specific language data poses a limitation on the applicability of these models. In this paper, we propose \ourmodel, a novel approach that combines automatic iterative prompting with program verification to ensure programs written in data-scarce domain-specific language are syntactically valid and incorporate environment constraints. Our approach provides effective guidance to the language model on generating structured-like task plans by incorporating any errors as feedback, while the verifier ensures the syntactic accuracy of the generated plans. We demonstrate the effectiveness of \ourmodel in planning chemistry experiments by achieving state-of-the-art results. We also show that the generated plans can be executed on a real robot by integrating them with a task and motion planner. 
\end{abstract}

\section{Introduction}

Leveraging natural language instruction to create a plan comes naturally to humans. However, when a robot is instructed to do a task, there is a communication barrier: the robot does not know how to convert the natural language instructions to lower-level actions it can execute, and the human cannot easily formulate lower-level actions.
Large language models (LLMs) can fill this gap by providing a rich repertoire of \textit{common sense reasoning} to robots~\cite{brown2020language, singh2022progprompt}.

Recently, there has been impressive progress in using LLMs~\cite{devlin2018bert, brown2020language,chowdhery2022palm} for problems involving structured outputs, including code generation~\cite{10.48550/arXiv.2107.03374, wang2021codet5, li2022competition} and robot programming~\cite{liang2022code}. These code generation models are often trained on code that is widely available on the Internet and perform well in few-shot settings for generating code in those languages.
However, to employ LLMs for task-plan generation there are two main issues to address:
(1)  \textit{lack of task-plan verification} and (2) poor \textit{performance for data-scarce domain-specific languages}.


\textbf{\paragraph{Lack of task plan verification}} Task plans generated by LLMs, often, cannot be executed out-of-the-box with robots. There are two reasons for that. First, machine-executable languages are bound by strict rules \cite{pddlstream}.
If the generated task plan does not adhere to them, it will not be executable. Hence, we need a way to verify the syntactic correctness of the structured task plan.
Second, LLMs might generate a task plan that looks reasonable (i.e. is syntactically correct) but is not actually executable by a robot. Avoiding this problem requires information about the world state and robot capabilities, as well as general reasoning about the physical world~\cite{brohan2022can}.



\textbf{\paragraph{Data scarcity for domain-specific languages}} It is difficult for LLMs to generate task plans in a zero-shot manner for domain-specific languages (DSLs), such as those in chemistry and physics because there is significantly less data on the Internet for those specific domains, so LLMs are unable to generalize well with no additional information \cite{gu2021domain, 10.48550/arXiv.2210.05359}. It is possible to address this by fine-tuning models on pairs of natural-language inputs and structured-language outputs, but it is very difficult to acquire training datasets large enough for the model to learn a DSL reasonably well~\cite{wang2021want}, and there is a large computation cost for fine-tuning LLMs \cite{bannour-2021-evaluating}. However, it has been shown that LLMs can adapt to new domains with \textit{effective prompting}~\cite{mishra2021cross}. Our insight is \textit{to leverage the in-context ability of an LLM by providing the rules of a structured language as input}, to generate a plan according to the template of the target DSL.

\begin{figure}[!t]
    \includegraphics[width=0.9\linewidth]{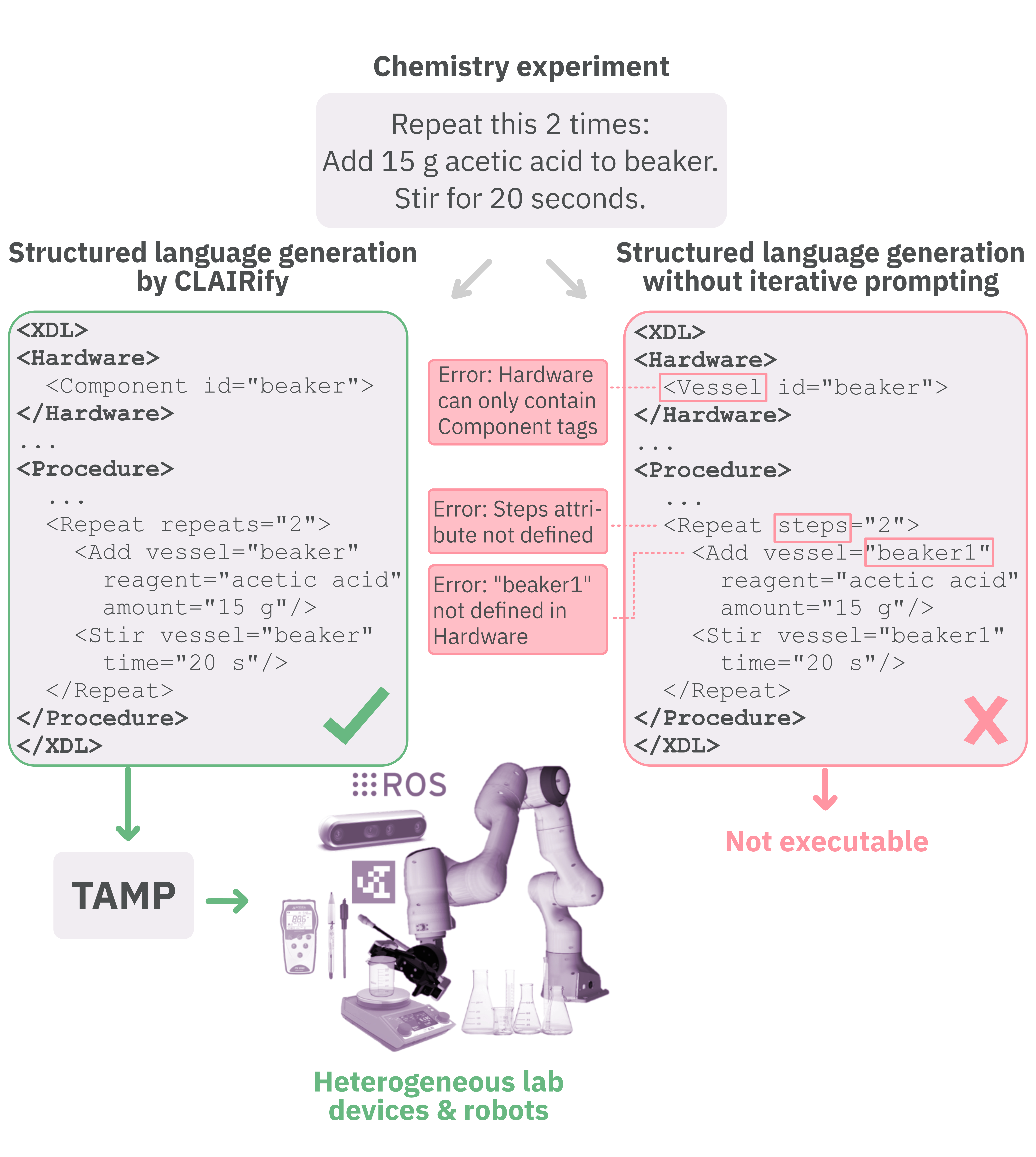}
    \centering
    \caption{Task plans generated by LLMs may contain syntactical errors in domain-specific languages. By using verifier-assited iterative prompting, \ourmodel can generate a valid program, which can be executed by a robot.}
    \label{fig:motivation}
\end{figure}

In this work, we propose to address the verification and data-scarcity challenges. We introduce \ourmodel\footnote{\textbf{\ourmodel website:} \href{https://ac-rad.github.io/clairify/}{https://ac-rad.github.io/clairify/}}, a framework that translates natural language into a domain-specific structured task plan using an automated iterative verification technique to ensure the plan is syntactically valid in the target DSL (Figure~\ref{fig:motivation}) by providing the LLM a description of the target language. Our model also takes into account environment constraints if provided.
The generated structured-language-like output is evaluated by our verifier, which checks for syntax correctness and for meeting environment constraints. The syntax and constraint errors are then fed back into the LLM generator to generate a new output. This iterative interaction between the generator and the verifier leads to grounded syntactically correct target language plans. 

We evaluate the capabilities of \ourmodel using a domain-specific language called Chemical Description Language (XDL)~\cite{10.1126/science.abc2986} as the target structured language unfamiliar to the LLM.
XDL is an XML-based DSL to describe action plans for chemistry experiments in a structured format, and can be used to command robots in self-driving laboratories~\cite{seifrid2022autonomous}. Converting experiment descriptions to a structured format is nontrivial due to the large variations in the language used. Our evaluations show that \ourmodel outperforms the current state-of-the-art XDL generation model in \cite{10.1126/science.abc2986}. We also demonstrate that the generated plans are executable by combining them with an integrated task and motion planning (TAMP) framework and running the corresponding experiments in the real world.
Our contributions are:

\begin{itemize}
    \item We propose a framework to produce task plans in a DSL using an iterative interaction of an LLM-based generator and a rule-based verifier.
    \item We show that the interaction between the generator and verifier improves zero-shot task plan generation.
    \item Our method outperforms the existing XDL generation method in an evaluation by human experts.
    \item We integrate our generated plans with a TAMP framework, and demonstrate  the successful translation of elementary chemistry experiments to a real robot execution.
\end{itemize}
\section{Related work}


\subsection{Task Planning} 
High-level task plans are often generated from a limited set of actions \cite{pddlstream}, because task planning becomes intractable as the number of actions and time horizon grows \cite{kaelbling2011hierarchical}.
One approach to do task planning is using rule-based methods~\cite{10.1126/science.abc2986, baier2009heuristic}. More recently, it has been shown that models can learn task plans from input task specifications \cite{sharma2021skill, mirchandani2021ella, shah2021value}, for example using hierarchical learning \cite{xu2018neural,huang2019neural}, regression based planning \cite{xu2019regression}, reinforcement learning \cite{eysenbach2019search}. However, to effectively plan task using learning-based techniques, large datasets are required that are hard to collect in many real-world domains. Our approach, on the other hand, generates a task plan directly from an LLM in a zero-shot way on a constrained set of tasks which are directly translatable to robot actions. We ensure that the plan is syntactically valid and meets environment constraints using iterative error checking. 

\subsection{Task Planning with Large Language Models}


Recently, many works have used LLMs to translate natural language prompts to robot task plans~\cite{brohan2022can, huang2022inner, liang2022code, singh2022progprompt}.  For example, Inner Monologue~\cite{huang2022inner} uses LLMs in conjunction with environment feedback from various perception models and state monitoring. However, because the system has no constraints, it can propose plans that are nonsensical. SayCan~\cite{brohan2022can}, on the other hand, grounds task plans generated by LLMs in the real world by providing a set of low-level skills the robot can choose from. A natural way of generating task plans is using code-writing LLMs because they are not open-ended (i.e. they have to generate code in a specific manner in order for it to be executable) and are able to generate policy logic. Several LLMs trained on public code are available, such as Codex~\cite{10.48550/arXiv.2107.03374},  CodeT5~\cite{wang2021codet5}, AlphaCode~\cite{li2022competition} and CodeRL~\cite{le2022coderl}. 
LLMs can be prompted in a zero-shot way to generate task plans. For example, Code as Policies~\cite{liang2022code} repurposes code-writing LLMs to write robot policy code and ProgPrompt~\cite{singh2022progprompt} generates plans that take into account the robot's current state and the task objectives. However, these methods generate Pythonic code, which is abundant on the Internet. For DSLs, naive zero-shot prompting is not enough; the prompt has to incorporate information about the target language so that the LLM can produce outputs according to its rules.

\subsection{Leveraging Language Models with External Knowledge}
A challenge with LLMs generating code is that the correctness of the code is not assured.
There have been many interesting works on combining language models with external tools to improve the reliability of the output.
Mind's Eye~\cite{10.48550/arXiv.2210.05359} attempts to ground large language model's reasoning with physical simulation. 
They trained LLM with pairs of language and codes and used the simulation results to prompt an LLM to answer general reasoning questions.
Toolformer~\cite{schick2023toolformer} incorporates API calls into the language model to improve a downstream task, such as question answering, by fine-tuning the model to learn how to call API. 
LEVER~\cite{10.48550/arXiv.2302.08468} improves LLM prompting for SQL generation by using a model-based verifier trained to verify the generated programs. 
As SQL is a common language, the language model is expected to understand its grammar. However, for DSLs, it is difficult to acquire training datasets 
and 
expensive to execute the plans to verify their correctness. 
Our method does not require fine-tuning any models or prior knowledge on the target language within the language model. Our idea is perhaps closest to LLM-A\textsc{ugmenter}~\cite{peng2023check}, which improves LLM outputs by giving it access to external knowledge and automatically revising prompts in natural language question-answering tasks. Our method similarly encodes external knowledge in the structure of the verifier and prompts, but for a structured and formally verifiable domain-specific language.



\begin{figure*}[!t]
    \begin{minipage}{0.8\linewidth}
    \centering
    \includegraphics[width=0.99\linewidth]{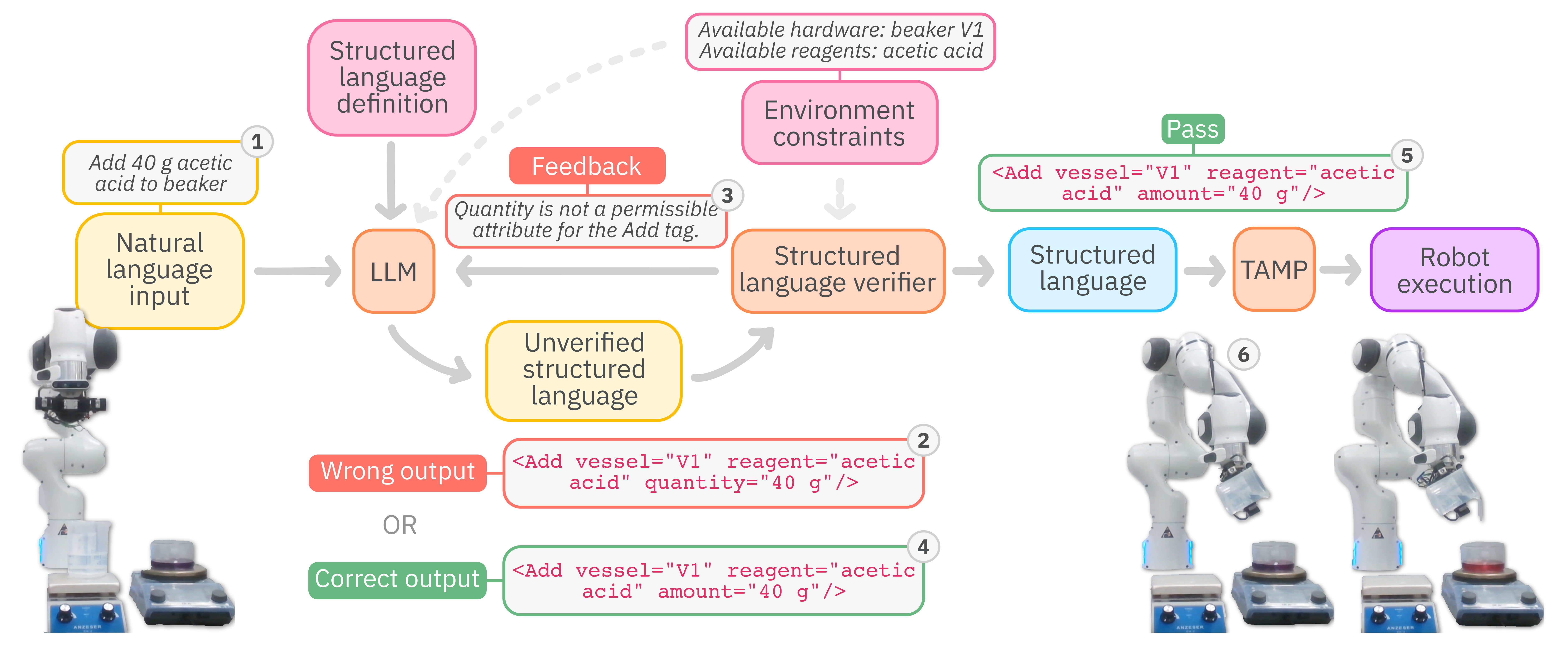}
    \end{minipage}
    \begin{minipage}{0.19\linewidth}
    \caption{\textbf{System overview}: The LLM takes the input (1), structured language definition, and (optionally) resource constraints and generates unverified structured language (2). The output is examined by the verifier, and is passed to LLM with feedback (3). The LLM-generated outputs passes through the verifier (4). The correct output (5) is passed to the task and motion planner to generate robot trajectories. The robot executes the planned trajectory (6).}
    \label{fig:system}
    \end{minipage}
\end{figure*}

\section{Task Programming with \ourmodel}


\paragraph{\ourmodel Overview}
We present a system that takes as input the specifications of a structured language (i.e. all its rules and permissible actions) as well as a task we want to execute written in natural language and outputs a syntactically correct task plan. 
A general overview of the \ourmodel pipeline is in Figure~\ref{fig:system}.
We combine input instruction and language description into a prompt and pass the prompt into the structured language generator (here we use GPT-3~\cite{brown2020language}, a large language model). However, we cannot guarantee the output from the generator is syntactically valid, meaning that it would definitely fail to compile into lower-level robot actions. To generate syntactically valid programs, we pass the output of the generator through a verifier. The verifier determines whether the generator output follows all the rules and specifications of the target structured language and can be compiled without errors. If it cannot, the verifier returns error messages stating where the error was found and what it was. This is then appended to the generator output and added to the prompt for the next iteration.
This process is repeated until a valid program is obtained, or until the timeout condition is reached (if that happens, an error is thrown for the user). Algorithm~\ref{alg:alg1} describes of our proposed method.

Once the generator output passes through the verifier with no errors, we are guaranteed that it is syntactically valid structured language. This program can then be translated into lower-level robot actions by passing through TAMP for robot execution. Each component of the pipeline is described in more detail below. 

\paragraph{\ourmodel in Chemistry Lab Automation} While our pipeline can in theory be applied to any structured language, we demonstrate it using the chemical description language (XDL)~\cite{10.1126/science.abc2986} as an example of a structured language. 
XDL describes the hardware and reagents to be used in the experiment, and the experimental procedures in chronological order.
Note that chemistry proceures are not well standardized; there is a lot of variation, and so translating them to a structured plan that can be executed by a robot is nontrivial~\cite{Vaucher2020}. 

\subsection{Generator}
The generator takes a user's instruction and generates structured-language-like output using a large language model (LLM) using a description of the structured language. The input prompt skeleton is shown in Snippet 1, Figure~\ref{fig:prompt}.
The description of the XDL language includes its file structure and lists of the available actions (can be thought of as functions), their allowed parameters and their documentation. 



Although the description of the target structured language information is provided, the candidate task plan is not guaranteed to be syntactically correct (hence we refer to is as ``structured-language-like'').
To ensure the syntactical correctness of the generated code, the generator is iteratively prompted by the automated interaction with the verifier.
The generated code is passed through the verifier, and if no errors are generated, then the code is syntactically correct.
If errors are generated, we re-prompt the LLM with the incorrect task plan from the previous iteration along with the list of errors indicating why the generated steps were incorrect.
The skeleton of the iterative prompt is shown in Snippet 2, Figure~\ref{fig:prompt}. 
Receiving the feedback from the verifier is used by the LLM to correct the errors from the previous iteration. This process is continued until the generated code is error-free or a timeout condition is reached, in which case we say we were not able to generate a task plan.

\begin{algorithm}[!t]
\footnotesize
\caption{{\footnotesize \ourmodel: Verifier-Assisted Iterative Prompts}}
\label{alg:alg1}
\begin{algorithmic}
\Require{Structured language description $\mathcal{L}$, instruction $x$}
\Ensure{Structured language task plan, $y_{SL}$}
\Procedure{IterativePrompting}{$\mathcal{L}$, $x$}
\State $y_{SL'} = \text{Generator}(\mathcal{L}, x)$
\State $ \text{errors} = \text{Verifier}(y_{SL'})$

\While{$\text{len(errors)} > 0 \text{ or timeout condition != True}$}
   \State $y_{SL'} = \text{Generator}(\mathcal{L}, x, y_{SL'}, \text{errors})$ 
   \State $ \text{errors} = \text{Verifier}(y_{SL'})$
   \EndWhile
\State $y_{SL} = y_{SL'}$
\EndProcedure
\end{algorithmic}
\end{algorithm}

\subsection{Verifier}
The verifier works as a syntax checker and static analyzer to  check the output of the generator and send feedback to the generator.
It first checks whether the input can be parsed as a correct XML and then checks the allowance of action tags, the existence of mandatory properties, and the correctness of optional properties.
This evaluates if the input is syntactically correct XDL.
It also checks the existence of definitions of hardware and reagents used in the procedure or provided as environment constraints, which works as a simple static analysis of necessary conditions for executability. If the verifier catches any of these errors, the candidate task plan is considered to be an invalid. The verifier returns a list of errors it found, which is then fed back to the generator.


\subsection{Incorporating Environment Constraints}
Because resources in a robot workspace are limited, we need to consider those constraints when generating task plans.
If specified, we include the available resources 
into the generator prompt.
The verifier also catches if the candidate plan uses any resources aside from those mentioned among the available robot resources.
Those errors are included in the generator prompt for the next iteration. 
If a constraint list is not provided, we assume the robot has access to all resources.
In the case of chemistry lab automation, those resources include experiment hardware and reagents. 

\subsection{Interfacing with Planner on Real Robot}
In many cases, the target DSL is not embodied, and it is hardware independent. Our verified task plan only contains high-level descriptions of actions.
To execute those actions by a robot, we need to map them to low-level actions and motion that the robot can execute.
To ensure the generated structured language is executable by the robot, we employ a task and motion planning (TAMP) framework. 
In our case, we use PDDLStream~\cite{pddlstream} to generate robot action and motion plans simultaneously.
In this process, visual perception information, coming from the robot camera, grounds the predicates for PDDLStream, and verified task plans are translated into problem definitions in PDDLStream.

For the chemistry lab automation domain, high-level actions in XDL are mapped to intermediate goals in PDDLStream, resulting in a long-horizon multistep planning problem definition.
To also incorporate safety considerations for robot execution, We use a constrained task and motion planning framework for lab automation~\cite{10.48550/arXiv.2212.09672} to execute the XDL generated by \ourmodel.

\begin{figure}[!t]
\begin{minipage}[t]{0.63\linewidth}
    \centering
\begin{lstlisting}[language=Python, caption=Initial prompt]
initial_prompt = """
# <Description of XDL>

# <Hardware constraints(optional)>
# <Reagent constraints (optional)>

Convert to XDL:
# <Natural language instruction>
"""
\end{lstlisting}

\begin{lstlisting}[language=Python, caption=Iterative prompt]
iterative_prompt = """
# <Description of XDL>

# <Hardware constraints(optional)>
# <Reagent constraints (optional)>

Convert to XDL:
# <Natural language instruction>
# <XDL from previous iteration>
This XDL was not correct.
There were the errors
# <List of errors, one per line>
Please fix the errors
"""
\end{lstlisting}
\end{minipage}
\hfill
\begin{minipage}[t]{0.35\linewidth}
\centering
\vspace{0.5em}
\caption{\textbf{Prompt skeleton}: (1) At the initial generation, we prompt the LLM with a description of XDL and the natural language instruction. (2) After the LLM generates structured-language-like output, we pass it through our verifier. If there are errors in the generated program, we concatenate the initial prompt with the XDL from the previous iteration and a list of the errors.}
    \label{fig:prompt}
\end{minipage}
\end{figure}


\section{Experiments and Evaluation}
Our experiments are designed to evaluate the following hypotheses: i) Automated iterative prompting increases the success rate of unfamiliar language generation, ii) The quality of generated task plans is better than existing methods, iii) Generated plans can be executed by actual hardware.



 \begin{table}[!t]
    \centering
    \caption{Comparison of our method with existing methods on the number of successfully generated valid XDL plans and their quality on 108 organic chemistry experiments from \cite{https://doi.org/10.5281/zenodo.3955107}.}
    \resizebox{\columnwidth}{!}{%
    \begin{tabular}{cccc}
        \hline
         Dataset & Method &  Number generated ↑ & Expert preference ↑ \\
         \hline\hline
         Chem-RnD & SynthReader \cite{10.1126/science.abc2986}   & 92/108 & 13/108\\
         & \ourmodel [ours] & \textbf{105/108} & 75/108 \\
         \hline
         Chem-EDU & SynthReader \cite{10.1126/science.abc2986}   & 0/40 & -\\
         & \ourmodel [ours] & \textbf{40/40} & - \\
         \hline
    \end{tabular}
    }
    \label{tab:comp_xdl}
\end{table}

\subsection{Experimental Setup}
To generate XDL plans, we use \texttt{text-davinci-003}, the most capable GPT-3 model at the time of writing. We chose to use this instead of \texttt{code-davinci-002} due to query and token limits. 

To execute the plans in the real world, we use an altered version of Franka Emika Panda arm robot, equipped with a Robotiq 2F-85 gripper, to handle vessels. The robot also communicates with instruments in the chemistry laboratory, such as a weighing scale and a magnetic stirrer. These devices are integrated to enable pouring and stirring skills.


\begin{figure*}[!t]
    \centering
    \includegraphics[width=0.63\linewidth]{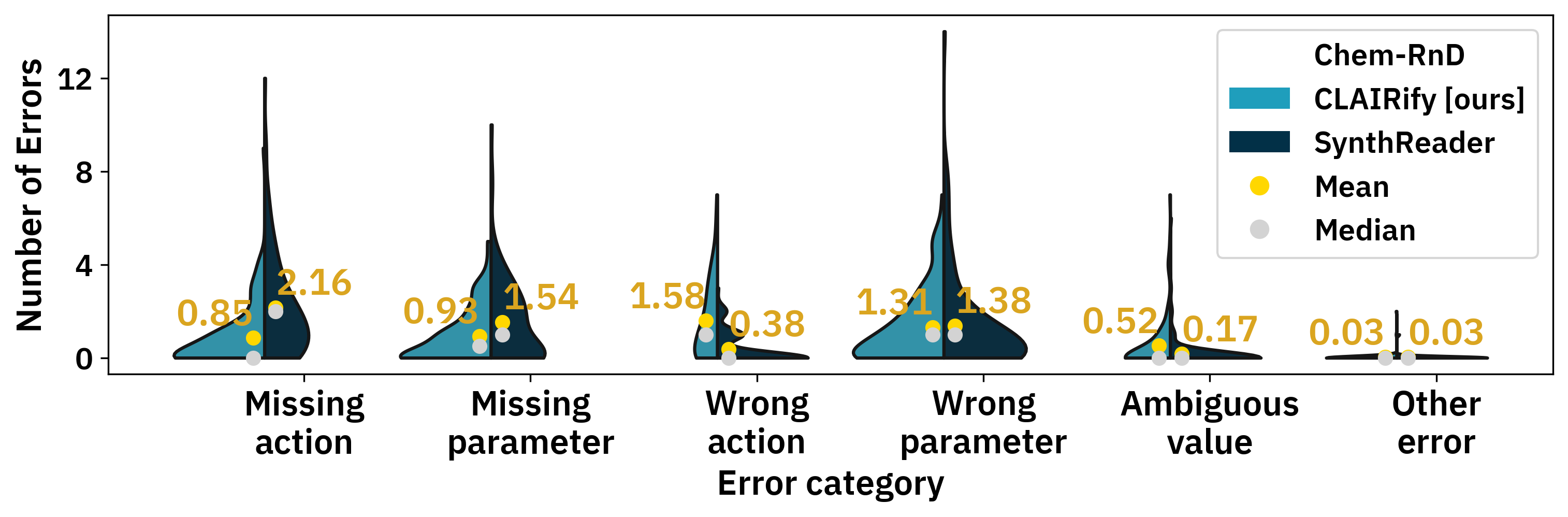}
    \includegraphics[width=0.36\linewidth]
    {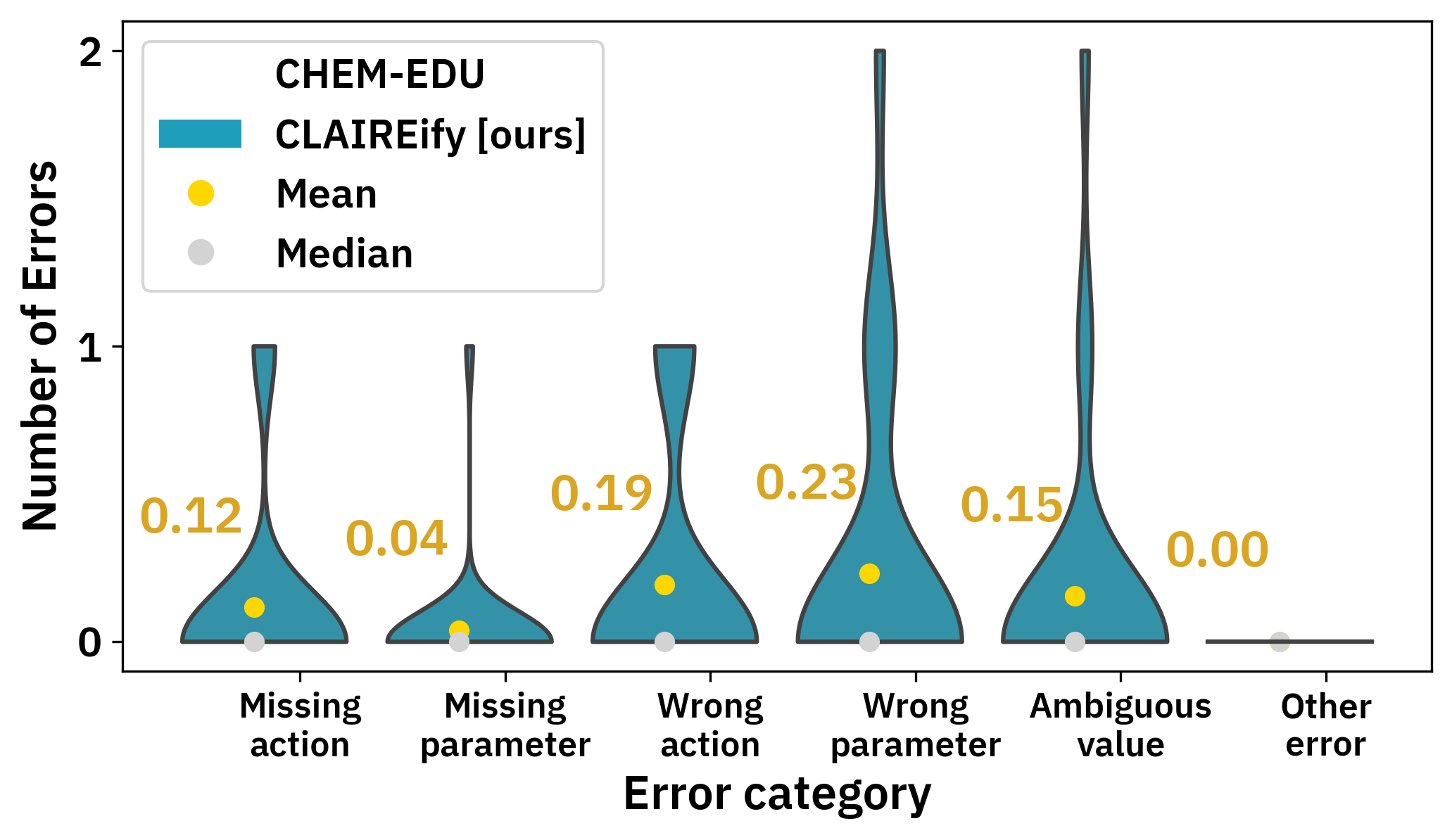}
    \caption{\textbf{Violin plots showing distributions of different error categories in XDL plans generated for experiments for the Chem-RnD (left) and Chem-EDU (right) datasets}. The x-axis shows the error categories and the y-axis shows the number of errors for that category (lower is better). For the Chem-RnD dataset, we show the error distributions for both \ourmodel and SynthReader. Each violin is split in two, with the left half showing the number of errors in plans generated from \ourmodel (teal) and the right half showing those from SynthReader (navy). For the Chem-EDU dataset, we only show the distributions for \ourmodel. In both plots, we show the mean of the distribution with a gold dot (and the number beside in gold) and the median with a grey dot.}
    \label{fig:error_breakdown_2}
\end{figure*}

\paragraph{\textbf{Datasets}} We evaluated our method on two different datasets: 

(1) \textbf{{Chem-RnD} [Chemistry Research \& Development]}:  This dataset consists of 108 detailed chemistry-protocols for synthesizing different organic compounds in real-world chemistry labs, sourced from the Organic Syntheses dataset (volume 77)~\cite{https://doi.org/10.5281/zenodo.3955107}. Due to GPT-3 token limits, we only use experiments with less than 1000 characters. We use Chem-RnD as a proof-of-concept that our method can generate task plans for complex chemistry methods. We do not aim to execute the plans in the real world, and so we do not include any constraints.

(2) \textbf{Chem-EDU [Everyday Educational Chemistry]}: We evaluate the integration of \ourmodel with real-world robots through a dataset of 40 natural language instructions containing only safe (edible) chemicals and that are, in principle, executable by our robot. The dataset consists of basic chemistry experiments involving edible household chemicals, including acid-base reactions and food preparation procedures\footnote{\textbf{\ourmodel Data \& code:} \href{https://github.com/ac-rad/xdl-generation/}{https://github.com/ac-rad/xdl-generation/}}.
When generating the XDL, we also included environment constraints based on what equipment our robot had access to (for example, our robot only had access to a mixing container called ``beaker"). 


\subsection{Metrics and Results}
The results section is organized based on the four performance metrics that we will consider, namely: Ability to generate structured-language output, Quality of the generated plans, Number of interventions required be the verifier, and Robotic validation capability. We compared the performance of our method with SynthReader, a state-of-the-art XDL generation algorithm which is based on rule-based tagging and grammar parsing of chemical procedures \cite{10.1126/science.abc2986}

    
    \indent (1) \textbf{Ability to generate a structured language plan.} First, we investigate the success probability for generating plans. For \ourmodel, if it is in the iteration loop for more than $x$ steps (here, we use $x=10$), we say that it is unable to generate a plan and we exit the program. When comparing with SynthReader, we consider that approach unable to generate a structured plan if the SynthReader IDE (called ChemIDE\footnote{ChemIDE using XDL: \href{https://croningroup.gitlab.io/chemputer/xdlapp/}{https://croningroup.gitlab.io/chemputer/xdlapp/}}) throws a fatal error when asked to create a plan. For both models, we also consider them unable to generate a plan if the generated plan only consists of empty XDL tags (i.e. no experimental protocol). For all experiments, we count the total number of successfully generated language plans divided by the total number of experiments. Using this methodology, we tested the ability of the two models to generate output on both the Chem-RnD and Chem-EDU datasets. The results for both models and both datasets are shown in Table ~\ref{tab:comp_xdl}. We find that out of 108 Chem-RnD experiments, \ourmodel successfully returned a plan 97\% of the time, while SynthReader returned a plan 85\% of the time. For the Chem-EDU dataset, \ourmodel generated a plan for all instructions. SynthReader was unable to generate any plans for that dataset, likely because the procedures are different from typical chemical procedures (they use simple action statements). This demonstrates the generalizability of our method: we can apply it to different language styles and domains and still obtain coherent  plans. 

    \indent (2) \textbf{Quality of the predicted plan (without executing the plan)}. To determine if the predicted task plans actually accomplish every step of their original instructions, we report the number of actions and parameters that do not align between the original and generated plan, as annotated by expert experimental chemists. To compare the quality of the generated plans between \ourmodel and SynthReader, we ask expert experimental chemists to, given two anonymized plans, either pick a preferred plan among them or classify them as equally good. We also ask them to annotate errors in the plans in the following categories:  Missing action, Missing parameter, Wrong action, Wrong parameter, Ambiguous value, Other error. Here, actions refer to high-level steps in the procedure (e.g., \texttt{<Add reagent="acetic acid">} is an action) and parameters refer to reagents, hardware, quantities and experiment descriptors (e.g., in \texttt{<HeatChill vessel="beaker" temp="100C">}, vessel and temp are both parameters). 
    The annotations were performed using the LightTag Text Annotation Tool \cite{perry-2021-lighttag}.

    \paragraph{Chem-RnD dataset}
    The results for the Chem-RnD dataset with respect to expert preference are reported in the last column of Table ~\ref{tab:comp_xdl}. We found that out of 108 experiments, experts preferred the XDL plan generated from \ourmodel 75 times and the one from SynthReader 13 times (the remaining 20 were considered to be of similar quality).

    The distributions of the annotated errors are shown in Figure~\ref{fig:error_breakdown_2}. We find that for 4 out of 6 error categories, our model does at least as well as or better than the baseline method when considering the mean and median of the distributions. We also find that for those categories, our method produces more experiments with 0 errors. 

    One advantage of our method is that it generates less plans with missing actions compared with the baseline. As XDL generation in SynthReader is implemented by rule-based pattern-matching techniques, any actions that do not match those templates would not appear in the final XDL. For example, for the protocol: 
\begin{lstlisting}[escapeinside={(*}{*)}]
To a solution of m-CPBA (200 mg, 0.8 mmol) in dichloromethane (10 mL), cooled to 0 (*$^\circ$*)C, was added dropwise a solution of 5-chloro-10-oxa-3-thia-tricyclo [5.2.1.01, 5] dec-8-ene (150 mg, 0.8 mmol) in dichloromethane (10 mL).
\end{lstlisting}
    
\noindent the plan generated by \ourmodel was
\begin{lstlisting}[language=xml, escapeinside={(*}{*)}]
<Add vessel="V1" reagent="m-CPBA" amount="200 mg"/>
<Add vessel="V1" reagent="dichloromethane" volume="10 mL"/>
<HeatChill vessel="V1" temp="0 (*\color{codepurple}{$^\circ$}*)C" time="3 min"/>
<Add vessel="V2" reagent="5-chloro-10-oxa-3-thia-tricyclo[5,2,1,0*1,5*]dec-8-ene" amount="150 mg"/>
<Add vessel="V2" reagent="dichloromethane" volume="10 mL"/>
<Transfer from_vessel="V2" to_vessel="V1"/>
\end{lstlisting}

\noindent while the plan generated from SynthReader was 
\begin{lstlisting}[language=xml]
<Add vessel="reactor" reagent="5-chloro-10-oxa-3-thia-tricyclo" volume="0" speed="40.0"/>
\end{lstlisting}

    Our model is able to decompose a complicated procedure into simpler actions by making two solutions in separate beakers and combining them with a Transfer procedure. It also assumes that the solutions don't already exist as mixtures and creates them from scratch. This is another benefit of our model, as it is able to understand implicit actions. For example, given the prompt

\begin{lstlisting}[]
L-Ornithine (31.92 g, 120 mmol) was added to a mixture of KOH (6.72 g, 120 mmol), water (200 ml) and THF (100 ml)
\end{lstlisting}

\noindent SynthReader considers a mixture of three chemicals as a single solution and creates the action:
\begin{lstlisting}[language=xml]
<Add vessel="reactor" reagent="a mixture of KOH (6.72 g, 120 mmol), water (200 ml) and THF (100 ml)" volume="0" speed="40.0"/>
<AddSolid vessel="reactor" reagent="L-Ornithine" mass="31.92 g"/>
\end{lstlisting}

    \noindent  On the other hand, \ourmodel correctly understand the implicit action to mix them beforehand and generates an appropriate XDL:

\begin{lstlisting}[language=xml]
<Add vessel="V1" reagent="L-Ornithine" amount="31.92 g"/>
<Add vessel="V1" reagent="KOH" amount="6.72 g"/>
<Add vessel="V1" reagent="Water" amount="200 ml"/>
<Add vessel="V1" reagent="THF" amount="100 ml"/>
\end{lstlisting}


    \begin{table}[!t]
        \centering
        \caption{\textbf{Verifier Analysis}. We report the average number of times \ourmodel calls the verifier for the experiments in a given dataset, as well as the minimum and maximum number of times. We also report the type of error encountered by the verifier and the number of times it caught that type. }
        \resizebox{\columnwidth}{!}{
        \begin{tabular}{cccp{5cm}}
            \hline
             Dataset &  Average num. & Max/min & Error type caught by verifier [count] \\
             &verifier calls&verifier calls& \\
             \hline \hline
             Chem-RnD  & $2.58 \pm 2.00$ & 10/1 & - missing property in action [306] \newline - property not allowed [174] \newline - wrong tag [120] \newline - action does not exist [21] \newline - item not defined in Hardware or Reagents list [15] \newline - plan cannot be parsed as XML [6] \\
             \hline
             \centering Chem-EDU & $1.15 \pm 0.45$ & 3/1 &  - item not defined in Hardware or Reagents list [47] \newline - property not allowed [26] \newline - wrong tag [40] \newline - missing property in action [3] \\
         \hline
        \end{tabular}
        }
        \label{tab:verifier_errors}
    \end{table}

    However, our model produced plans with a greater number of wrong actions than SynthReader. This is likely because our model is missing domain knowledge on certain actions that need to be included in the prompt or verifier. For example, given the instruction \textit{"Dry solution over magnesium sulfate"}, our model inserts a \texttt{<Dry .../>} into the XDL plan, but the instruction is actually referring to a procedure where ones passes the solution through a short cartridge containing magnesium sulphate, which seems to be encoded in SynthReader. Another wrong action our model performs is reusing vessels. In chemistry, one needs to ensure a vessel is uncontaminated before using it. However, our model generates plans that can use the same vessel in two different steps without washing it in between. 
    Our model also sometimes generates plans with ambiguous values. For example, many experiment descriptions include conditional statements such as ``Heat the solution at the boiling point until it becomes white''. Conditions in XDL need a numerical condition as a parameter. Our model tries to incorporate them by including actions such as \texttt{<HeatChill temp="boiling point" time="until it becomes white"/>}, but they are ambiguous. We can make our model better in the future by incorporating more domain knowledge into our structured language description and improving our verifier with real-world constraints. For example, we can incorporate visual feedback from the environment, include look-up tables for common boiling points, and ensure vessels are not reused before cleaning.

    Despite the XDL plans generated by our method containing errors, we found that the experts placed greater emphasis on missing actions than ambiguous or wrong actions when picking the preferred output, 
    indicating larger severity of this class of error for the tasks and outputs investigated here.

    \begin{table}[!t]
        \centering
        \caption{Number of XDL plans successfully generated for different error message designs in the iterative prompting scheme on a validation set from Chem-RnD.}
        \resizebox{\columnwidth}{!}{%
        \begin{tabular}{p{7.5cm}p{2.5cm}}
            \toprule
              Variations of Iterative Prompt Design using Verifier Error Messages &  Plan's generated success rate (\%) ↑\\
             \hline \hline
             \textit{Naive}:  XDL from previous iteration and string “This XDL was not correct. Please fix the errors.” &  0 \\
             \hline
             \textit{Last Error}: Error List from verifier from 
             previous iteration  & 30\\
             \hline
             \textit{All Errors cumulative}: Accumulated error List from all previous iterations & 50\\
             \hline
             \textit{XDL + Last Error}: XDL and Error List from verifier from previous iteration & 100 \\
             \bottomrule
        \end{tabular}
        }
        \label{tab:ablation}
    \end{table}

    \begin{figure*}[!t]
        \centering
        \includegraphics[width=0.85\linewidth]{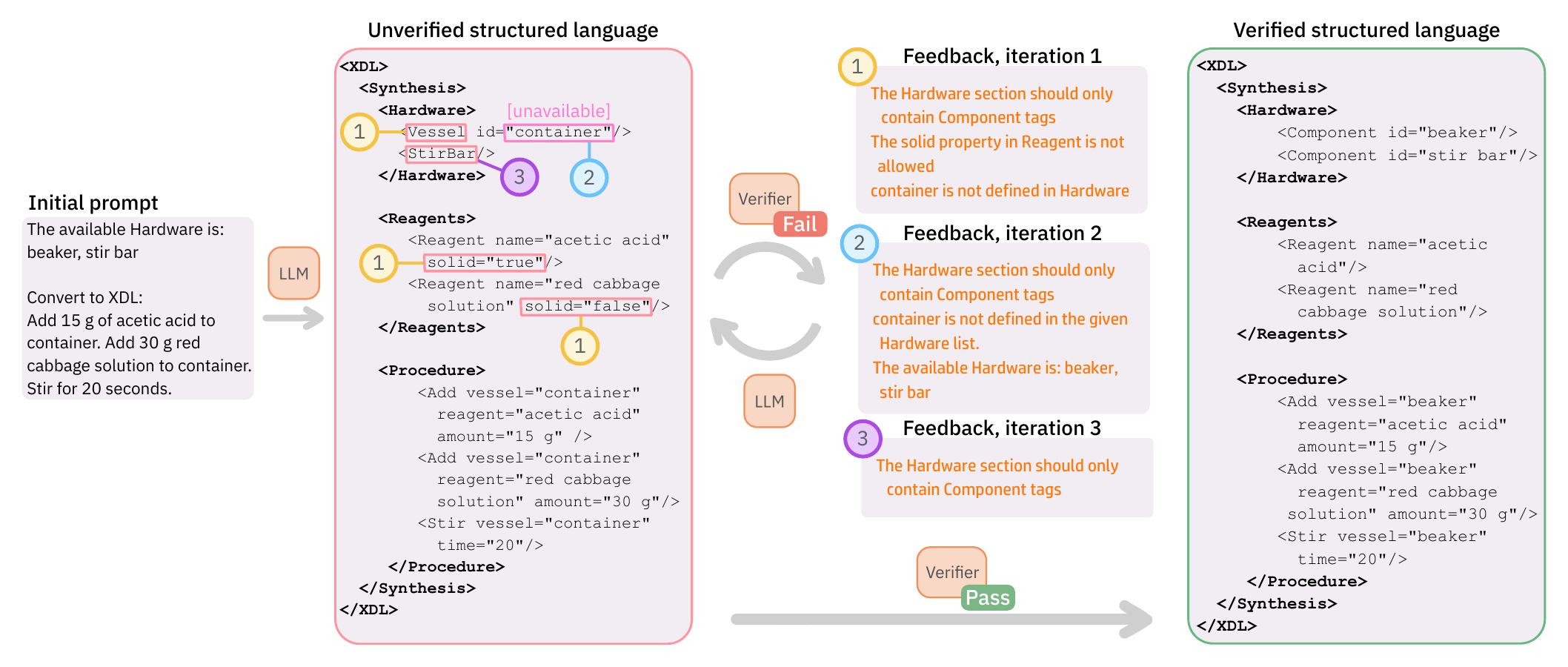}
     \caption{\textbf{Feedback loop between the Generator and Verifier.} The input text is converted to structured-like language via the generator and is then passed through the verifier. The verifier returns a list of errors (marked with a yellow 1). The feedback is passed back to the generator along with the erroneous task plan, generating a new task plan. Now that previous errors were fixed and the tags could be processed, new errors were found (including a constraint error that the plan uses a vessel not in the environment). These errors  are denoted with a blue 2. This feedback loop is repeated until no more errors are caught, which in this case required 3 iterations.}
        \label{fig:iter_loop}
    \end{figure*}

    \paragraph{Chem-EDU dataset}
    We annotated the errors in the Chem-EDU datasets using the same annotation labels as for the Chem-RnD dataset. The breakdown of the errors is in the left plot of Figure~\ref{fig:error_breakdown_2}. Note that we did not perform a comparison with SynthReader as no plans were generated from it. 
    We find that the error breakdown is similar to that from Chem-RnD, where we see amibiguous values in experiments that have conditionals instead of precise values. We also encounter a few wrong parameter errors, where the model does not include units for measurements. This can be fixed in future work by improving the verifier to check for these constraints. 

    \indent (3)  \textbf{Number of interventions required by the verifier.} To better understand the interactions between the generator and verifier in \ourmodel, we analyzed the number of interactions that occur between the verifier and generator for each dataset to understand the usefulness of the verifier. In Table~\ref{tab:verifier_errors}, we show that each experiment in the Chem-RnD dataset runs through the verifier on average 2.6 times, while the Chem-EDU dataset experiments runs through it 1.15 times on average. The difference between the two datasets likely exists because the Chem-EDU experiments are shorter and less complicated. The top Chem-EDU error encountered by the verifier was that an item in the plan was not define in the Hardware or Reagents list, mainly because we included hardware constraints for this dataset that we needed to match in our plan. In Figure~\ref{fig:iter_loop}, we show a sample loop series between the generator and verifier. 
    
    \indent (4) \textbf{Robotic validation (Chem-EDU only).} To analyze how well our system performs in the real world, we execute a few experiments from the Chem-EDU dataset on our robot. Three experiments from the Chem-EDU dataset were selected to be executed.

    \paragraph{Solution Color Change Based on pH}
    As a basic chemistry experiment, we demonstrated the color change of a solution containing red cabbage juice. This is a popular introductory demonstration in chemistry education, as the anthocyanin pigment in red cabbage can be used as a pH indicator~\cite{10.1021/ed069p66.1}.
    We prepared red cabbage solution by boiling red cabbage leaves in hot water. The colour of the solution is dark purple/red. Red cabbage juice changes its color to bright pink if we add an acid and to blue if we add a base, and so we acquired commercially-available vinegar (acetic acid, an acid) and baking soda (sodium bicarbonate, a base).

    In this experiment, we generated XDL plans using \ourmodel from two language inputs: 
    \vspace{-5pt}
\begin{lstlisting}[escapeinside={(*}{*)}]
[1] Add 40 g of red cabbage solution into a beaker. Add 10 g of acetic acid into the beaker, then stir the solution for 10 seconds.
\end{lstlisting}
\vspace{-10pt}
\begin{lstlisting}[escapeinside={(*}{*)}]
[2] Add 40 g of red cabbage solution into a beaker. Add 10 g of baking soda into the beaker, then stir the solution for 10 seconds.
\end{lstlisting}
\vspace{-5pt}

\begin{figure*}[!t]
    \begin{minipage}{0.75\linewidth}
    \centering
    \includegraphics[width=\textwidth]{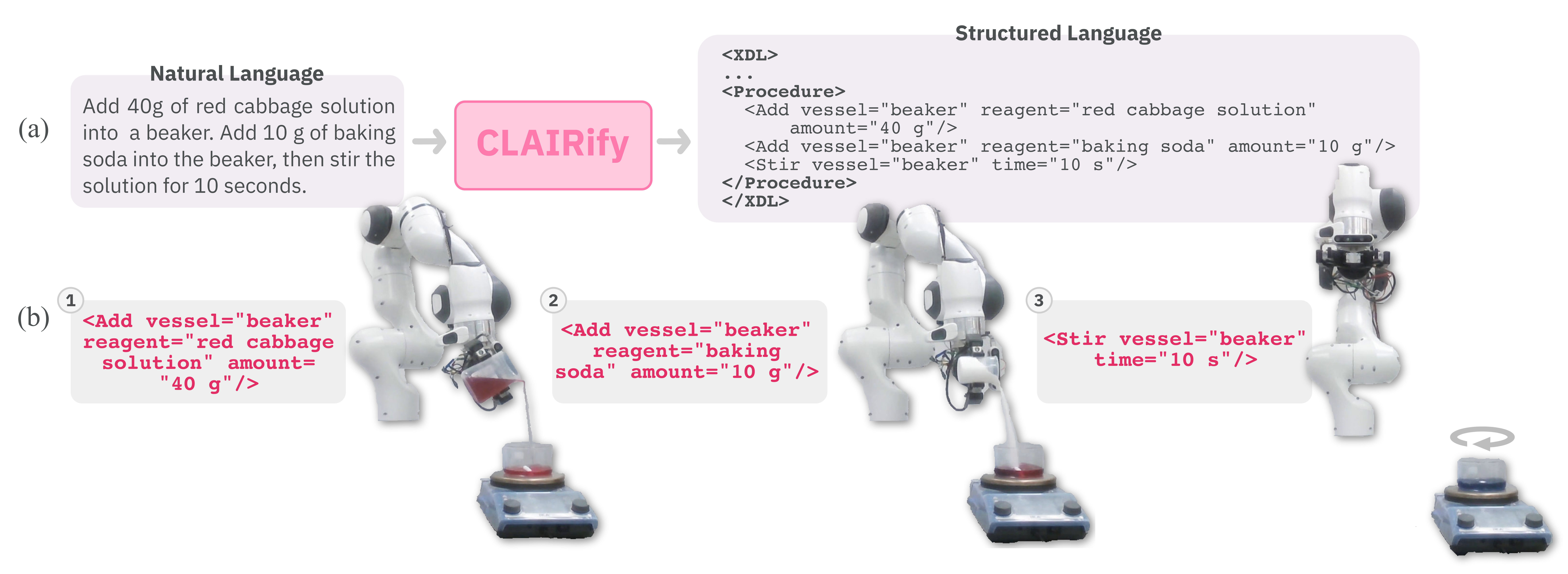}
    \end{minipage}
    \begin{minipage}{0.24\linewidth}
    \caption{\textbf{Robot execution}: The robot executes the motion plan generated from the XDL for given natural language input. (a) \ourmodel converts the natural language input from the user into XDL. (b) The robot interprets XDL and performs the experiment. Stirring is done by a rotating stir bar inside the beaker.
    }    \label{fig:robot_setup}
    \end{minipage}
\end{figure*}

Figure~\ref{fig:robot_setup} shows the flow of the experiment. Our generated a XDL plan that correctly captured the experiment; the plan was then passed through TAMP to generate a low-level action plan and was then executed by the robot. 

\paragraph{Kitchen Chemistry}
We then tested whether our robot could execute a plan generated by our model for a different application of household chemistry: food preparation.
We generated a plan using \ourmodel for the following lemonade beverage, which can be viewed on our website:

\begin{lstlisting}[escapeinside={(*}{*)}]
Add 15 g of lemon juice and sugar mixture to a cup containing 30 g of sparkling water. Stir vigorously for 20 sec.
\end{lstlisting}
\vspace{-5pt}

\subsection{Ablation Studies} 
We assess the impact of various components in our prompt designs and feedback messaging from the verifier. We  performed these tests on a small validation set of 10 chemistry experiments from Chem-RnD (not used in the test set) and report the number of XDL plans successfully generated (i.e., was not in the iteration loop for $x=10$ steps).

\paragraph{\textbf{Prompt Design}}
To evaluate the prior knowledge of the GPT-3 on XDL, we first tried prompting the generator without a XDL description, i.e., with the input: 

\begin{lstlisting}[language=Python]
initial_prompt = """
Convert to XDL:
# <Natural language instruction>"""
\end{lstlisting}

The LLM was unable to generate XDL for any of the inputs from the small validation set that contains 10 chemistry experiments. For most experiments, when asked to generated XDL, the model output a rephrased version of the natural language input. In the best case, it output some notion of structure in the form of S-expressions or XML tags, but the outputs were very far away from correct XDL and were not related to chemistry. We tried the same experiment with \texttt{code-davinci-002}; the outputs generally had more structure but were still nonsensical. 
This result suggests the LLM does not have the knowledge of the target language and including the language description in the prompt is essential to generate an unfamiliar language. 

\paragraph{\textbf{Feedback Design}}
We experimented with prompts in our iterative prompting scheme containing various levels of detail about the errors. The baseline prompt contains a description as well as natural language instruction. We wanted to investigate how much detail is needed in the error message for the generator to be able to fix the errors in the next iteration. For example, is it sufficient to write ``There was an error in the generated XDL", or do we need to include a list of errors from the verifier (such as ``Quantity is not a permissible attribute for the Add tag"), or do we also need to include the erroneous XDL from the previous iteration? 

We find that including the erroneous XDL from the previous iteration and saying why it was wrong resulted in the high number of successfully generated XDL plans. Including a list of errors was better than only writing ``This XDL was not correct. Please fix the rrors", which was not informative enough to fix any errors. Including the erroneous XDL from the previous iteration is also important; we found that including only a list of the errors without the context of the XDL plan resulted in low success rates.

\section{Conclusion and Future Work}
In this paper, we introduce \ourmodel to generate structured language task plans in a DSL by providing an LLM with a description of the language in a zero-shot manner. We also ensure that the task plan is syntactically correct in the DSL by using a verifier and iterative prompting. Finally, we show that our plans can incorporate environmental constraints. We evaluated the performance of \ourmodel on two datasets and find that our method was able to generate better plans than existing baselines. Finally, we translate a select number of these plans to real-world robot demonstrations. 

In the future, we will incorporate feedback from the robot planner and environment into our plan generation process. We will also improve the verifier to encode knowledge of the target domain nuances.
With these technical refinements, we expect our system can become integrated even better in robotic task planning for different domains. 

\section{Acknowledgements}
We would like to thank members of the Matter Lab for annotating task plans. We would also like to thank the Acceleration Consortium for their generous support, as well as the Carlsberg Foundation. 
\renewcommand*{\bibfont}{\small}
\bibliographystyle{IEEEtran}
\bibliography{reference}
\end{document}